\documentclass[10pt,twocolumn,letterpaper]{article}

\usepackage{cvpr}
\usepackage{times}
\usepackage{epsfig}
\usepackage{graphicx}
\usepackage{amsmath}
\usepackage{amssymb}
\usepackage{algorithm}
\usepackage{algorithmic}
\usepackage{multirow}

\usepackage[breaklinks=true,bookmarks=false]{hyperref}

\cvprfinalcopy % *** Uncomment this line for the final submission

\def\cvprPaperID{****} % *** Enter the CVPR Paper ID here

\ifcvprfinal\fi
\begin{document}

%%%%%%%%% TITLE
\title{Learning adaptively from the unknown for few-example video person re-ID}

\author{Jian Han\\
Tsinghua University\\
{\tt\small hanj19@mails.tsinghua.edu.cn}
}

%\and
%Takashi Isobe \\
%Tsinghua University\\
%{\tt\small jbj18@mails.tsinghua.edu.cn}

\maketitle
%\thispagestyle{empty}

%%%%%%%%% ABSTRACT
\begin{abstract}
This paper mainly studies one-example and few-example video person re-identification. A multi-branch network PAM that jointly learns local and global features is proposed. PAM has high accuracy, few parameters and converges fast, which is suitable for few-example person re-identification. We iteratively estimates labels for unlabeled samples, incorporates them into training sets, and trains a more robust network. We propose the static relative distance sampling(SRD) strategy based on the relative distance between classes. For the problem that SRD can not use all unlabeled samples, we propose adaptive relative distance sampling (ARD) strategy. For one-example setting, We get 89.78\%, 56.13\% rank-1 accuracy on PRID2011 and iLIDS-VID respectively, and 85.16\%, 45.36\% mAP on DukeMTMC and MARS respectively, which exceeds the previous methods by large margin.
\end{abstract}

%%%%%%%%% BODY TEXT
\section{Introduction}
Person re-identification aims at spotting the same person from non-overlapping  camera views, which can be applied to crime suspect recognition, target customer identification and other scenarios.

\begin{figure}[t]
\begin{center}
\includegraphics[width=1.0\linewidth]{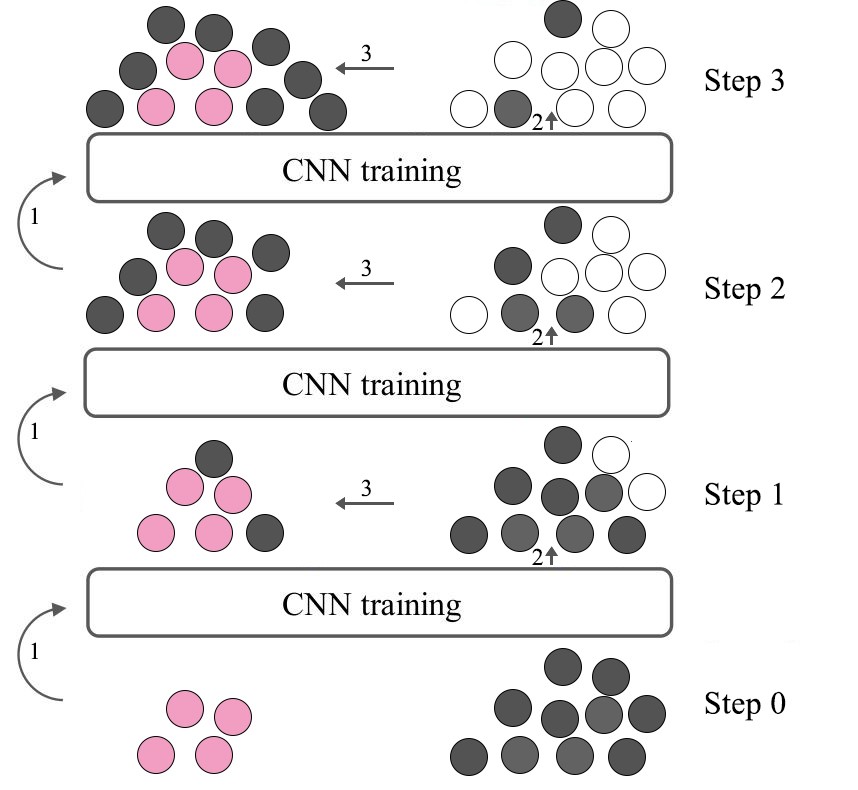}
\end{center}
   \caption{An illustration of the proposed method's procedure. We first initialize the CNN with the labeled samples(1). And then the initialized CNN is used to embed the unlabeled samples into the feature space. We assign the closest labeled sample's label to the unlabeled(2). Then we use adaptive relative distance sampling strategy to select reliable ones and join them into training set(3). We retrain the CNN use the new training set(1). We iterate the whole procedure until all unlabeled samples have joined the training set.}
\label{fig:long}
\label{fig:onecol}
\end{figure}

As a key technology in intelligent video surveillance, person re-identification has received great attention of scholars and lots of excellent results have emerged. Especially since the introduction of the deep learning method, the feature extraction and metric learning are integrated together. More
discriminative feature representation and distance metric can be learned, far exceeding the traditional two-stage method of hand-crafted features and separate distance metric learning.
\begin{figure*}[t]
\begin{center}
\includegraphics[width=0.9\linewidth]{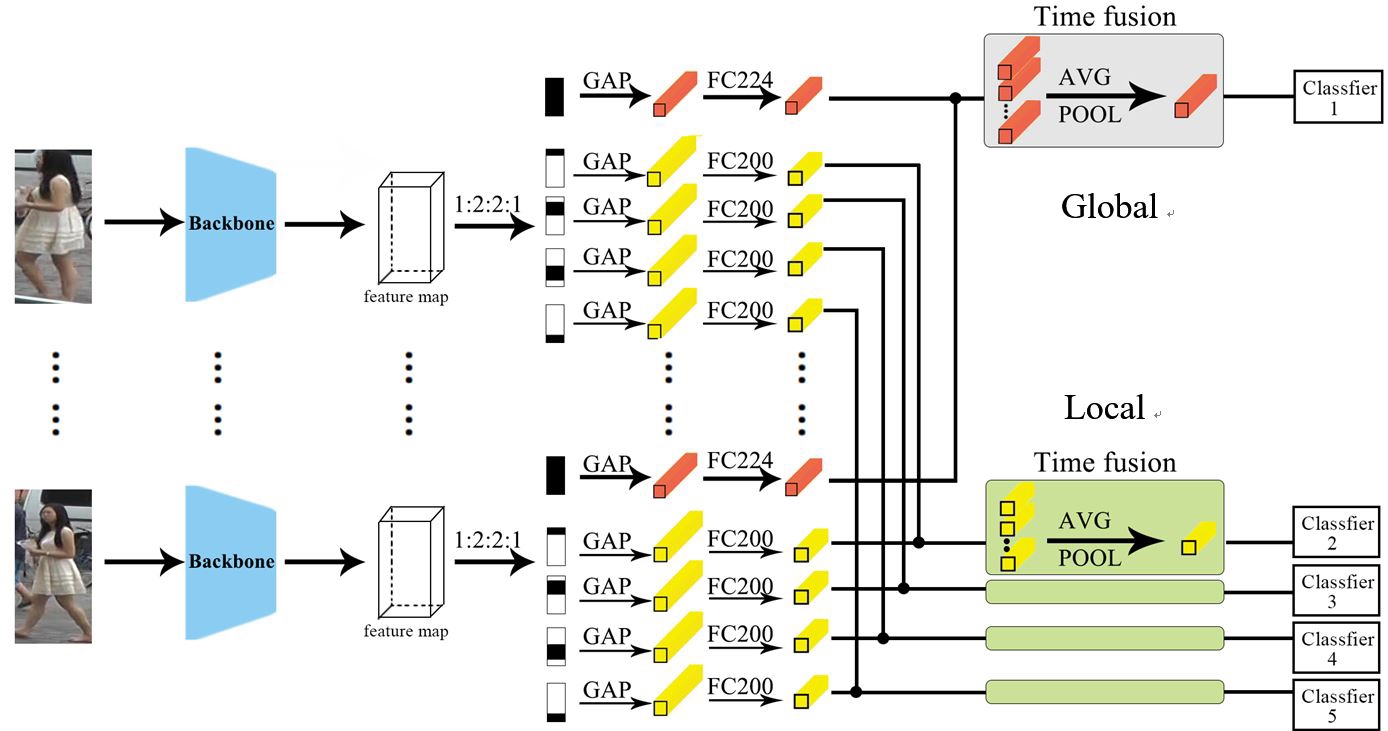}
\end{center}
   \caption{The framework of Part Attention Model. We first feed T frames of a tracklet into backbone and get T feature maps. The upper global branch first exert global average poolling(GAP) and then dimension reduction. After time fusion, we get the global featrue. Four parts cut vertically from the feature map are fed into four local branches. After that we get four local feature.}
\label{fig:long}
\label{fig:onecol}
\end{figure*}
Although person re-identification has made great progress, there are still some problems. Mainly in the following two aspects: First, the existing deep learning method is dominated by a fully supervised network, relying on a large number of annotated data. In practical applications, the real situation of the scene is complex and changes with time. Obtaining a large number of annotation data suitable for the application scene is very expensive and impractical, which limits the application of the existing supervised algorithm relying on a large number of annotation data. Second, the existing research algorithms have achieved good results on a limited set of general data sets, but when applied to actual scenes, they are subject to a series of complexities such as occlusion, illumination changes, pedestrian attitude changes, viewing angle changes, and even camera model changes. The impact of factors will greatly reduce the accuracy of recognition. Therefore, the existing research is still far from the actual application level.

It's very expensive and impractical to rely on a large amount of manual labeling data in practical applications. But a large number of unlabeled samples can be obtained from person detection and tracking in surveillance video. Few-example learning method only needs to label the sample with a small label or a small amount. By correctly estimating the label for the unlabeled sample, it is used to train the network, which can solve the problem that the manual labeling is expensive and the time consuming is too long in the practical application. Therefore, few-example person re-identification method has great research value and practical value.

As illustrated in Figure 1, we iteratively estimates labels for unlabeled samples, incorporates them into training sets, and trains a more robust network.

Our contributions can be summarized as follows:
\begin{itemize}
    \item A multi-branch network PAM that jointly learns local and global features is proposed. PAM has high accuracy, few parameters and converges fast, which is suitable for few-example person re-identification.
    \item We propose the static relative distance sampling(SRD) strategy based on the relative distance between classes which surpass GPS on small-scale datasets. For the problem that SRD can not use all unlabeled samples, we propose adaptive relative distance sampling (ARD) strategy.
    \item For one-example experiment, PAM+GPS reaches 86.9\% rank-1 accuracy and 47.26\% mAP on DukeMTMC-VID and MARS respectively. PAM+ARD reaches 89.78\%, 56.13\%, 89.17\% rank-1 accuracy and 45.36\% mAP on PRID2011, iLIDS-VID, DukeMTMC-VID and MARS respectively, which exceeds the state-of-the-art by a large margin.
\end{itemize}

\section{Related work}
In recent years, approaches based on deep neural networks have dominated the research of person re-identification. These approaches combined the feature representation learning and distance metric learning together with an end-to-end architecture.

\textbf{Supervised Person Re-ID.}
Dong Yi et al. \cite{yi_deep_2014} introduce siamese network to person re-identification and used cosine similarity to measure similarity loss. Although siamese network has many excellent features, it's easy to cause the network training failure because the cross entropy loss is very sensitive to small changes in the feature vector. In order to solve this problem, S. Ding et al. \cite{ding_deep_2015} used a triplet composed of positive and negative sample pairs as the input of the siamese network. By learning pairs of positive and negative samples, feature embeedings become more discriminative. At the same time, siamese network and common classification network are combined in their work, while using the Softmax loss and the improved triple loss. Zhao et al. \cite{zhao_person_2017} introduce saliency information into person re-identification, learning the saliency of pedestrians and matching the significant similarities of different pedestrians. W. Li et al. [8] propose a multi-branch network. The upstream single branch learns the global feature of pedestrians, and the downstream multi-branches learn multiple local features. Meanwhile, the interaction between soft and hard space, channel attention and spatial attention is used. The network proposed by S. Li et al. \cite{li_harmonious_2018} can learn multiple spatial partial attention models, and adopts a diversity regularity term to ensure that multiple partial attention models focus on different areas of the body. The time-attention mechanism employed enables the network to learn the features of the face, torso, and other parts of the body from the best-conditioned frames in the sequence.

\textbf{Few-example person Re-ID.}
In general, there are two main types of few-example methods. One is to establish a good model only with a small number of labeled data. These methods may use siamese network, matching network, Meta Learning, and Transfer Learning. The other is to train a good network by estimating the label for the unlabeled data and then augmenting the training set with them.

In the first type, Koch et al. \cite{koch_siamese_nodate} propose a framework for solving the problem of one-shot classification. They first build a fully convolutional siamese network based on verification loss, and then use this network to calculate the similarity between the image to be identified and other labeled samples. The image is then recognized as a sample of the category which the most similar labeled sample belongs to. Vinyals et al. \cite{vinyals_matching_2016} propose matching network. During the training process, some samples are selected to form a support set and the remaining samples are used as training images. They construct different encoders for the support set and training pictures. The classfier's output is a weighted sum of the predicted values between the support set and the training images. During the test process, one-shot sample are used as support set to predict the category of new images. Rahimpour et al. \cite{rahimpour_attention-based_2018} use meta-learning methods to learn multiple similar tasks, and build two encoders for the gallery and probe respectively. Based on these encoders, they get gallery images' embedding according to the characteristics of the remaining gallery images. They get probe images' embedding according to the characteristics of the gallery images. In this way they obtain a more discriminative feature representation.

In the second type, Ye et al. \cite{ye_dynamic_2017} establish a graph for each camera. They view the labeled sample as the node of the graph, and view the distance between the video sequence features as the path. Unlabeled sample are mapped into different graphs (namely estimating the labels) to minimize the objective function. The graphs are updated dynamically . They continually estimate labels, and train models until the algorithm converges. Liu et al. \cite{liu_stepwise_2017} first initialize the model with labeled samples. Then they calculate k nearest neighbors of the probe with the gallery. They remove the suspect samples and then add the remaining samples to the training set. The procedure is iterated until the algorithm converges. Wu et al. \cite{wu_exploit_2018} initialize a CNN with labeled data firstly, and then linearly incorporate pseudo-label samples to the training set according to the distance to labeled samples. Then the CNN is retrained with the new training set. Finally all unlabeled samples have estimated label and are added into training set, then they use a validation set to select the best model.

\begin{figure}[t]
\begin{center}
\includegraphics[width=0.9\linewidth]{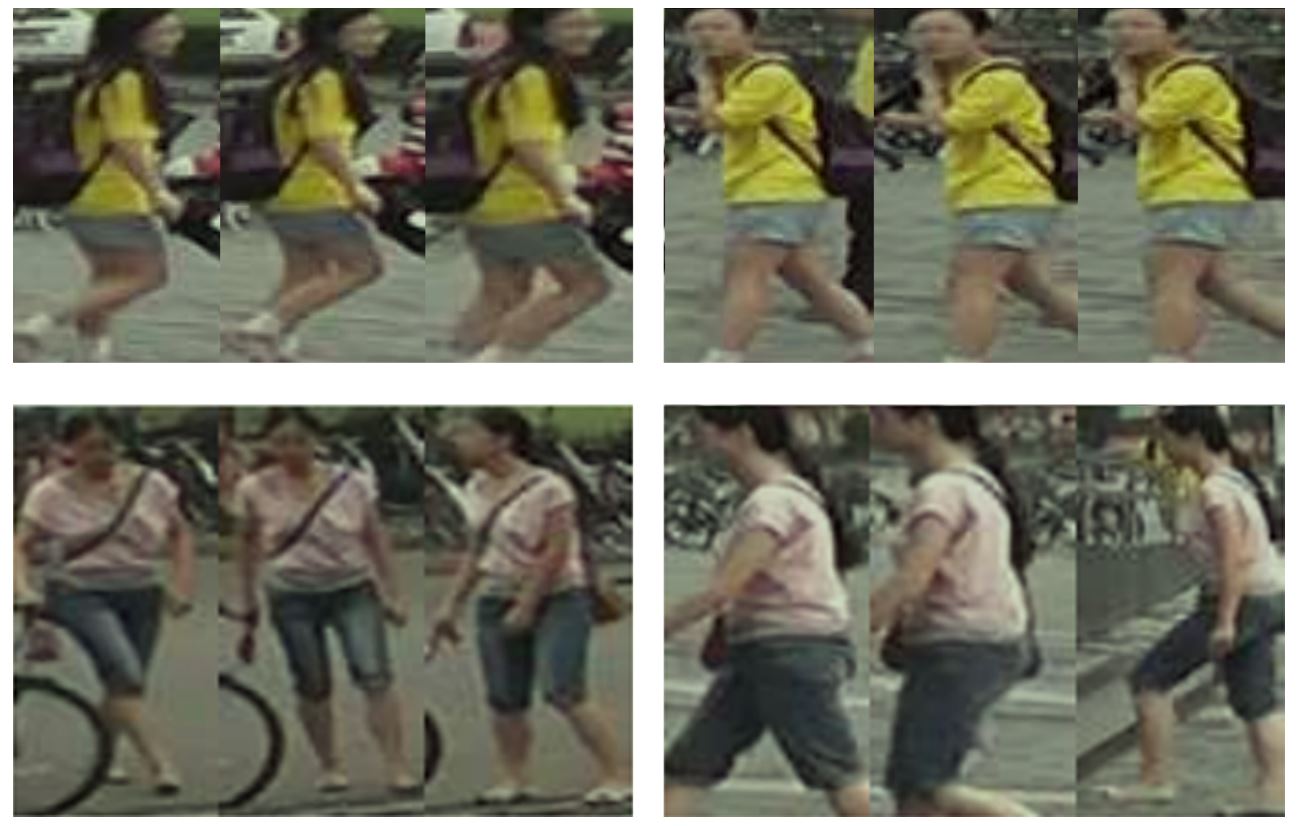}
\end{center}
   \caption{A false identification example of absolute distance sampling. The absolute distance of the first row which blongs to different person is smaller than the second row which blongs to the same person.}
\label{fig:long}
\label{fig:onecol}
\end{figure}

\section{Method}
\subsection{Framework Overview}
The framework of our approach is shown in Figure 1. We first initialize the CNN with the labeled samples. The CNN here is the PAM network (section 3.2). Then we use the trained CNN to estimate the labels for the unlabeled samples based on the distance between the labeled samples and unlabled samples. The label of the labeled samples closest to the unlabeled sample in the feature space is used as the estimated label of the unlabeled sample. Unlabeled samples and their estimated labels form pseudo-label samples. Then we use sampling strategy ARD(section 3.3) to select correctly estimated ones and then incorporate them into training set. After that, the enlarged training set is used to re-train the CNN again. We iterate this process until all unlabeled samples have been estimated and added to the training set. Since the training set is enlarged continuously during training iterations, we can progressively learn a more stable model. When the algorithm converges, we get a most robust model trained with all samples.

\begin{algorithm}[htb]
\caption{PAM+ARD}
\label{alg:Framwork}
\begin{algorithmic}[1] %这个1 表示每一行都显示数字
\REQUIRE ~~\\ %算法的输入参数：Input
Labeled data $L_{-} d a t a=\left\{\left(x_{1}, y_{1}\right),\ldots,\left(x_{N}, y_{N}\right)\right\}$\\
Unlabled data $U_{-} d a t a=\left\{\tilde{x}_{1}, \tilde{x}_{2}, \ldots, \tilde{x}_{M}\right\}$\\
CNN model $\Phi\left(\bullet ; \theta_{0}\right)$ \\
\ENSURE ~~\\
Best CNN model $\Phi\left(\bullet ; \theta^{*}\right)$ \\
\STATE Initialize CNN model $\Phi\left(\bullet ; \theta_{0}\right)$ with $L_{-}data$ \\
\label{ code:fram:Initialize }%对此行的标记，方便在文中引用算法的某个步骤
\STATE Estimate labels $U_{-}data$ $\rightarrow\left\{\left(\tilde{x}_{1}, \tilde{y}_{1}\right),\ldots,\left(\tilde{x}_{M}, \tilde{y}_{M}\right)\right\}$\\
\label{code:fram:Estimate}
\STATE $k=k_{0}$
\label{code:fram:add}
\STATE \textbf{do}
\label{code:fram:classify}
\STATE $~~k=k_{0}+0.1$
\label{code:fram:add}
\STATE ~~Sample $D_{\text {intra}}<k \times D_{\text {inter}}$ from $U_{-}$ data $\rightarrow P_{-}$ data
\label{code:fram:add}
\STATE \textbf{while} $| P_{-}$ data $\left|<0.15 \times \right| L_{-}$ data $|$
\label{code:fram:add}
\STATE $k_{s}=k$
\label{code:fram:add}
\STATE \textbf{for} $k=k_{s} \rightarrow 1.0$ \textbf{do}
\label{code:fram:add}
\STATE ~~\textbf{while} $\left|P_{-} d a t a_{k t}\right|-\left|P_{-} d a t a_{k-1}\right|$ \\
~~~~~~~~~~~~$<(-k) \times \left|P_{-} d a t a_{k 1}\right|-\left|P_{-} d a t a_{k 0}\right|$ ~\textbf{do}
\label{code:fram:add}
\STATE ~~~~$Train_{-}data = \left\{L_{-}data, P_{-}data\right\} $
\label{code:fram:add}
\STATE ~~~~Re-train the CNN model $\Phi\left(\cdot ; \theta_{0}\right)$ with $Train_{-}data$
\label{code:fram:add}
\STATE ~~~~Est. labels $U_{-}data$ $\rightarrow\left\{\left(\tilde{x}_{1}, \tilde{y}_{1}\right),\ldots,\left(\tilde{x}_{M}, \tilde{y}_{M}\right)\right\}$
\label{code:fram:add}
\STATE ~~~~Sam. $D_{\text {intra}}<k \times D_{\text {inter}}$ from $U_{-}data$ $\rightarrow P_{-}data_{kt}$
\label{code:fram:add}
\STATE ~~\textbf{end while}
\label{code:fram:add}
\STATE \textbf{end for}
\label{code:fram:add}
\RETURN $\Phi\left(\bullet ; \theta^{*}\right)$ %算法的返回值
\end{algorithmic}
\end{algorithm}

\subsection{Part Attention Model}
In this paper, the small sample method of iteratively estimating labels is adopted. If the model is too complex and the number of parameters is too large, the iteration time is too long. If the model performance is not good, the label estimation accuracy is low, so that all the training sets are error samples, which will lead to poor algorithm performance. Therefore, the network must be simple in structure and high in accuracy to meet the needs. Most of the existing fully supervised network models with good performance are complex, with large number of parameters and slow convergence, which cannot be applied in this paper. However, the network performance with few parameters can not meet the demand. To solve this problem, this paper proposes a multi-branch network for joint learning of local and global features.

The network structure is shown in Figure 2, using a multi-branch network that learns both local and global features. A tracklet is randomly extracted from the frame and input into the network. After Backbone, a feature map is obtained. Then, the feature map is sent to the global branch as a whole, and the feature matrix of the dimension is obtained through global pooling and dimension reduction. Then, the global feature of the dimension is obtained by merging the time domain average pooling in the time domain; the feature graph is sliced vertically. After obtaining p local feature maps, the local feature maps are globally pooled and dimensionally reduced to obtain p-dimensional feature matrices, and then the time-domain averaged pools are combined in the time domain to obtain p-dimensional local features.

Backbone uses the Resnet50 pre-trained on ImageNet\cite{krizhevsky_imagenet_2012}. The Global branch is the original IDE network [18], and the Local branch uses the 1:2:2:1 slice for the feature map output of the last pooled layer of ResNet50. The global feature and each local feature each train a classifier. Enter a tracklet to extract the global features and 4 local features of each frame, and concatenate the global feature and four local features as the feature representation of the entire tracklet.
\subsection{Adaptively Sampling Distance Strategy}
The task of the sampling strategy is to select all the correctly estimated pseudo-label samples to join the training set as much as possible. If the pseudo-label samples sampled after each iteration are correct, but the number is small, then adding it to the training set and training CNN again can bring very limited performance improvement. The next time the tag is estimated, it will only be the last time. The result is slightly better, and the increase in the number of correct labels that can be selected will be very slow, resulting in a network that has been growing in performance, but the training takes too long. On the other hand, if the number of pseudo-label samples selected by each iteration is large, but it is very correct, adding it to the training set to train the CNN will result in a decrease in network performance and loss of the meaning of increasing the number of training samples. Therefore, the number of pseudo-label samples selected by the sampling strategy is required to be large and the accuracy is high.

Wu et al. [15] adopted a linearly increasing sampling strategy. After each iteration, the nearest neighbor-based tag estimation strategy is used to estimate the tag for the unlabeled sample, and then all the unlabeled samples are represented by the L2 distance between the feature space and the nearest tagged sample to represent the credibility of the tag estimate. It sorts from small to large, and then samples where t is the number of iterations and p is the tunable parameter), and has achieved good performance on the MARS and DukeMTMC-VID data sets. However, there are some problems with this sampling method. For a limited number of unlabeled samples, label estimation is difficult. Each iteration adds a fixed number of samples than the last time, causing the initial iterations to include only simple samples that are easy to estimate, wasting the network's ability to estimate. The last few iterations added too many difficult samples to estimate the error, overdrafting the network capabilities. In order to estimate and sample more robustly, the author set the parameter p to 0.05 in the experimental setup, requiring a total of 20 iterations, and the training time is too long. Moreover, the selection of the p value is a very difficult problem. A smaller p value can bring better performance, but the iteration time is too long. Larger p account for severe performance degradations due to the addition of too many false samples.

The setting of the small sample determines that there are only a small number of samples when the network is initialized, and the network tends to learn the simple and direct distinction between the samples, such as the color of the clothes, and ignores other higher-level distinguishing information. If the sampling only depends on the absolute distance between the feature space samples, the shallow information will be similar and the distance between the samples of the actual different classes (the first line in Figure 4-1) is less than the shallow information is not so similar but actually the same kind. The distance between the samples (the second line in Figure 4-1) is preferentially added to the network. The result is catastrophic, because any unlabeled sample with similar surface information (wearing a yellow T-shirt) will be estimated to be added to the training set due to the smaller absolute distance, thus forming a malignant Positive feedback damages the discriminating power of the network. This is also the root cause of poor performance of PAM plus GPS algorithm on one example of small-scale data set PRID2011, iLIDS-VID.
\subsubsection{Static Relative Distance Sampling}
We propose a Static Relative Distance Sampling (SRD) based on the distance relationship between classes: The distance between the unlabeled sample in the feature space and the nearest labeled sample is the label sample and the rest of the label. However, the minimum distance between samples whose labels are not (ie, not belonging to the same class) is that when the a equation is satisfied, the unlabeled sample and its estimated label are added to the training set. Iteratively trains the network, estimates the label, and samples the extended training set. When the difference between the number of samples and the number of previous samples is less than b (for the manually set hyperparameter, it can be set to 0.01, 0.03, etc.) the algorithm converges and stops iterating.

The reason for this is that CNN can learn a good feature extractor and distance measurement, and it can distinguish between similar samples and different samples after embedding samples into the feature space. If the estimated tag of the unlabeled sample is correct, the distance between the unlabeled sample and the nearest labeled sample belongs to the intra-class distance. However, the distance between samples with tags and other samples with tags but different (that is, they do not belong to the same class) is inter-class distance, among which the smallest distance is the smallest inter-class distance. When the intra-class distance is less than the minimum inter-class distance, we have reason to believe that the estimated tag is correct, so we add it to the training set. In order to improve the accuracy, I multiplied the minimum inter-class distance by a constant k less than 1 to expand the training set more carefully. Obviously, the smaller the k, the higher the probability that the sample you add has the correct label.

SRD static sampling can achieve good performance, but there are two problems: first, although the convergence can be achieved after iterating a few steps, the convergence cannot use up all the unlabeled samples. Second, as the number of iterations increases, the number of samples in the last few iterations increases very little, but it takes up a lot of training time. We hope that the algorithm can make use of all the unlabeled samples and add more unlabeled samples for most purposes in each iteration, so as to save too many iterations and reduce meaningless learning.

Based on the static sampling strategy SRD, we propose an Adaptive Relative Distance Sampling (ARD) based on the distance between classes. The core idea is to find the appropriate initial k value through the probe mechanism, and then perform SRD. When the relative value of the current iterative sampling number increases is less than a certain threshold, k is automatically increased, and the SRD is performed under the new k value, so that more samples are added in the next iteration. Use all unlabeled samples until k>1.
The adaptive sampling strategy (ARD) specifically includes a k-probe mechanism and an adaptive k-value increasing mechanism, which are explained in detail below.

\textbf{k-probe mechanism..}
The goal of the k-probe mechanism is to find a suitable k-value to start sampling. Considering that the initial k value will increase the k value after the convergence of the algorithm, it will continue to iterate. Therefore, the k value is suitable from a small value. At the same time, considering that the k value is too small, there may be no qualified pseudo-label samples, or join. The number of pseudo-label samples is too small, making meaningless iterations, wasting the predictive power of the network. Therefore, the following k-probe mechanism is proposed: CNN estimates the label for the unlabeled sample after initializing the training with the labeled sample. Then try to sample k 0.6, 0.7, 0.8, 0.9, 1.0 in sequence, and stop when the number of samples is greater than the initial k value.

\textbf{Adaptively increase k value mechanism.}
For each k, record the difference $k_{-}margin0$ of the number of samples of the previous two iterations of the k value. If the difference between the current iteration sample number and the last sample number is less than (-k) $\times$ $k_{-}margin0$, then k=k + 0.1, increase The k value continues to train the network. When k exceeds 1, the training is terminated and the algorithm converges. The threshold (-k) $\times$ $k_{-}margin0$ takes a dynamic setting method, setting different thresholds for different k, and small k sets a larger threshold. The function of setting the threshold for each k value is to determine whether the SRD sample under the current k value has converged. When the increased number of samples is less than the threshold, the value of k is increased to start the next SRD, and the purpose of adaptive sampling is achieved. $k_{-}margin0$ is determined by the difference between the number of samples of the previous two iterations of the current k value, and is different for different k. This threshold setting method is a great innovation in this paper and is the core of the Adaptive Sampling Strategy (ARD).
\section{Experiments}

\begin{table*}[]
\begin{center}
\begin{tabular}{l|ll|ll|ll|ll}
\hline
\multirow{2}{*}{Methods} & \multicolumn{2}{c|}{PRID2011}                      & \multicolumn{2}{c|}{iLIDS-VID}                     & \multicolumn{2}{c|}{DukeMTMC}                      & \multicolumn{2}{c}{MARS}                         \\ \cline{2-9}
                         & \multicolumn{1}{c}{Rank-1} & \multicolumn{1}{c|}{mAP} & \multicolumn{1}{c}{Rank-1} & \multicolumn{1}{c|}{mAP} & \multicolumn{1}{c}{Rank-1} & \multicolumn{1}{c|}{mAP} & \multicolumn{1}{c}{Rank-1} & \multicolumn{1}{c}{mAP} \\ \hline
EUG  (CVPR18)            & 59.6                    & 65.3                     & --                      & --                       & 72.79                   & 63.23                    & 62.67                   & 42.45                   \\
DGM  (ICCV17)            & 82.4                    & --                       & 37.1                    & --                       & 44.36                   & 33.62                    & 36.81                   & 16.87                   \\
Stepwise(ICCV17)         & 84.27                   & 87.64                    & --                      & --                       & 56.26                   & 46.76                    & 41                   & 19.65                   \\
BUC\cite{Lin2019ABC}(AAAI19)              & --                      & --                       & --                      & --                       & 69.2                    & 61.9                     & 61.1                    & 38.0                    \\
TAUDL\cite{ferrari_beyond_2018}(ECCV18)            & 49.4                    & --                       & 26.7                    & --                       & --                      & --                       & 43.8                    & 29.1                    \\
PAM+SRD(Alg2.)           & 85.51                   & 87.60                    & 39.80                   & 45.59                    & 83.33                   & 77.13                    & 56.82                   & 38.71                   \\
PAM+ARD(Alg3.)           & 89.78                   & 94                    & 56.13                   & 61.14                    & 89.17                   & 85.16                    & 61.57                   & 45.36                   \\ \hline
\end{tabular}
\end{center}
\caption{ Comparison with the state-of-the-art methods on iLIDS-VID, PRID2011, DukeMTMC-VID, and MARS. All the methods are conducted based on one-example setting except BUC. Although DGM, Stepwise, TAUDL claim that their methods are unsupervised. Their methods belongs to one-example methods strictly. }
\end{table*}

\subsection{Datasets and Settings}
\textbf{The iLIDS-VID dataset.}
The iLIDS-VID dataset \cite{fleet_person_2014} contains 300 different pedestrians shot by two non-overlapping cameras with a total of 600 tracklets (each tracker has 2 tracklets). Each tracklet has a length of 23-192 frames and an average length of 73 frames. Because this data set was taken in the multi-camera network in the arrival hall of the airport, the clothing similarity, illumination, and viewing angle vary greatly, so it is more challenging.

\textbf{The PRID2011 dataset.}
The PRID2011 dataset \cite{heyden_person_2011} contains 934 pedestrians shot by two still cameras with different perspectives, including a total of 1134 tracklets. Camera 1 took 385 tracklets from 385 pedestrians, Camera 2 shot 749 tracklets from 749 pedestrians, and only the first 200 pedestrians appeared in both cameras. Each tracklet has a length of 5-675 frames and an average length of 100 frames.

\textbf{The DukeMTMC dataset.}
The DukeMTMC-VID dataset \cite{wu_exploit_2018} is a large-scale video person re-ID dataset that is processed by the DukeMTMC picture pedestrian re-identification data set. It is photographed by a plurality of cameras whose fields of view do not overlap, and is manually labeled by constructing a tracklet by equally extracting 12 frames per second in continuous video. A total of 1404 pedestrian tracklets (two tracklets for each pedestrian with at least two different cameras) and 408 pedestrian jamming tracklets (one tracklet for each pedestrian with only one camera), a total of 4832 tracklets. 2196 tracklets of 702 pedestrians in the data set were used for training, and 2,636 tracklets of the remaining 702 pedestrians and 408 disturbing pedestrians were used for testing.

\textbf{The MARS dataset.}
The MARS dataset \cite{noauthor_mars:_nodate} is the largest data set for video pedestrian recognition, and is expanded by the Market1501 data set. Shot on a college campus by six near-synchronized cameras, with 1,261 segments of 1,631 segments of tracklets and 3,248 segments of interference tracklets (error detection or tracking video sequences). It is divided into a training set of 625 pedestrians and a test set of 636 pedestrians. Each pedestrian has an average of 13 tracklets, 816 frames, and each pedestrian has at least two tracklets taken by different cameras. Another significant feature of the MARS dataset and the above dataset is that it uses algorithmic annotation rather than manual annotation. The detection and tracking of pedestrian bounding boxes uses the Variable Part Model (DPM) [19] and many more. Target tracking algorithm GMMCP tracker [20].

\textbf{Experiment Setting.}For one-example experiments, we use the same protocol as [21]. In both datasets, we randomly choose one tracklet in camera 1 for each identity as initialization. If there is no tracklet recorded by camera 1 for one identity, we randomly select one tracklet in the next camera to make sure each identity has one video tracklet for initialization. For few-example experiments, we conduct it on the MARS dataset. 20\%, 40\%, and 60\% of the samples were randomly selected from the training set as the initial labeled data. The remaining samples of the training set are stripped of its labels and used as unlabeled samples.

\textbf{Implementation Details.}
Train 70 epochs on the iLIDS-VID and PRID2011 data sets, using the random gradient descent (SGD) plus momentum (Momentum) optimization method, the momentum is set to 0.5, the initial initial learning law is set to 0.1, 55 epochs Set to 0.01 afterwards.
50 epochs were trained in the DukeMTMC-VID and MARS data sets, and the random gradient plus momentum optimization method was also used. The momentum was set to 0.5, the initial learning law was set to 0.1, and the learning law was set to 0.01 after 40 epochs. .
The weight attenuation is set to 5e-4; the data enhancement method using random cropping, random flipping, and random erasure: random cropping to size, random horizontal flip with a probability of 0.5, and random erase area area ratio range [0.02, 0.2], an area with an aspect ratio of [0.3, 3.3], filled with pixels [0.0, 0.0, 0.0].
Fixed the parameters of conv1, layer1 and layer2 of ResNet50. The learning rate of the rest of ResNet50 is set to 0.1 times of the global learning rate; the value of the loss function is set to 0.1, and the value of K is the number of categories classified in the training stage (that is, the number of ids of the training set) ).

\subsection{Comparison with the State-of-the-Art Methods}

From the experimental results of the four datasets, the algorithm 2 static sampling SRD and the algorithm 3 adaptive sampling ARD have good performance. Especially on the small-scale datasets PRID2011 and iLIDS-VID, the performance of one-example exceeds the state of the art and is better than the PAM+GPS of Algorithm 1. Because it is based on the distance between the classes and the relative distance between the classes, more metric information is used, which can overcome the problems of GPS equalization incremental sampling in section 4.1.1, so it is obtained on PRID2011 and iLIDS-VID. A significant performance boost.
The SRD algorithm of static sampling is not as good as the GPS algorithm on the DukeMTMC and MARS datasets, mainly because GPS can estimate more more in multiple iterations by using a smaller n0 and a smaller sampling increase factor k. The correct label. However, when the SRD algorithm is set too small, it will fall into local optimum on the big data set, and only a small number of unlabeled samples are added. The excessive k setting will cause too many error samples to be added in the initial iteration, which will gradually deteriorate as the number of iterations increases, which limits the performance improvement. After adaptive incremental sampling ARD, the performance of DukeMTMC and MARS data sets has been significantly improved, which proves the effectiveness of adaptive incremental sampling strategy compared to pure static sampling strategy, especially on large-scale data sets.
Adaptive Sampling ARD (Algorithm 3) exceeds the previous algorithm on all four data sets. Rank-1 was 89.78\% and 56.13\% on PRID and iLIDS-VID, respectively, and 85.16\% and 45.36\% on DukeMTMC and MARS, respectively, although the MARS data set was slightly inferior to the PAM+GPS algorithm proposed in Chapter 3. However, the performance is better on small-scale data sets, and the number of iterations is also lower, so the overall performance is the best.
\subsection{Few-example experiment}
The results of the few-example settings on the MARS dataset using PAM and the Adaptive Sampling Algorithm (ARD) are shown in Table 1.

\begin{table}[]
\begin{center}
\begin{tabular}{l|l|l|l|l}
\hline
\multicolumn{1}{c|}{No.} & \multicolumn{1}{c|}{Method} & \multicolumn{1}{c|}{Type} & \multicolumn{2}{c}{MARS} \\ \cline{4-5}
                         &                           &                           & R-1         & mAP         \\ \hline
1                        & AMOC+EpicFlow\cite{liu_video-based_2016}             & Super.                    & 68.3        & 52.9        \\
2                        & QAN\cite{liu_quality_2017}                       & Super.                    & 73.7        & 51.7        \\
3                        & PAM+ARD                   & Super.                    & 61.57       & 45.36       \\
4                        & PAM+ARD                   & Semi.(20\%)               & 68.38       & 52.61       \\
5                        & PAM+ARD                   & Semi.(40\%)               & 74.29       & 60.31       \\
6                        & PAM+ARD                   & Semi.(60\%)               & 77.98       & 65.74       \\ \hline
\end{tabular}
\end{center}
\caption{Comparison between the few-example performance of our method and some supervised methods. The percentage in the bracket indicates the ratio of labeled samples.}
\end{table}

As can be seen from the above 1, the PAM+SRD algorithm achieves 52.61\% when using 20\% of the labeled data, which is better than the full-supervised algorithm QAN, which is slightly inferior to the full-supervised algorithm AMOC+EpicFlow. Explain that our method can achieve the performance of the fully supervised algorithm when only 20\% of the labeled data is used, which further proves the excellent performance of the algorithm. Although few-example setting requires more manual labeling than single labeling, performance can be greatly improved.

\subsection{Ablation Studies}
We performed a series of ablation experiments on the PAM+ARD algorithm on the DukeMTMC-VID dataset according to one example to verify the performance of each part of the algorithm.

% \usepackage{multirow}
\begin{table}[]
\begin{center}
\begin{tabular}{l|lllll}
\hline
\multirow{2}{*}{Methods} & \multicolumn{5}{c}{DukeMTMC-VID}                                                                                     \\ \cline{2-6}
                         & \multicolumn{1}{l|}{R-1}  & \multicolumn{1}{l|}{R-5}  & \multicolumn{1}{l|}{R-10} & \multicolumn{1}{l|}{R-20} & mAP  \\ \hline
IDE+ARD           & \multicolumn{1}{l|}{x} & \multicolumn{1}{l|}{x} & \multicolumn{1}{l|}{x} & \multicolumn{1}{l|}{x} & x \\
PAM+EUG(k=0.05)   & \multicolumn{1}{l|}{x} & \multicolumn{1}{l|}{x} & \multicolumn{1}{l|}{x} & \multicolumn{1}{l|}{x} & x \\
PAM+ARD           & \multicolumn{1}{l|}{89.2} & \multicolumn{1}{l|}{96.7} & \multicolumn{1}{l|}{97.9} & \multicolumn{1}{l|}{98.3} & 85.2 \\ \hline
\end{tabular}
\end{center}
\caption{The ablation studies of our method. In the first experiment, we replace the CNN with common IDE network. In the second experiment, we replace the sampling strategy with linear growth sampling strategy proposed in \cite{wu_exploit_2018}.}
\end{table}

\textbf{Part attation model.}
Table 3 is a control variable experiment for the Part Attention Model (PAM). The algorithm uses the Adaptive Increased Sampling Strategy, and the CNN networks are the IDE and PAM. The results show that PAM can achieve a 2.7\% performance improvement over Rank on the DukeMTMC-VID dataset, which demonstrates the effectiveness of Part Attention Model.

\textbf{Sampling strategy.}
Table 3 is a control variable experiment for the Adaptive Increased Sampling Strategy (ARD). The algorithm uses the IDE network, and the sampling strategy uses the linear increase sampling (p=0.05) and adaptive incremental sampling (ARD) of the EUG. The results show that ARD can achieve a 2.7\% performance improvement over Rank on the DukeMTMC-VID dataset, which demonstrates the effectiveness of adaptively increasing the sampling strategy.

\textbf{Dynamic coefficient for threshold.}
In order to verify the effectiveness of the method of setting thresholds by multiplying kmargin0 of different k values with different coefficients (-k), I designed a group of comparative experiments: one group multiplied kmargin0 of different k values with dynamic coefficients (-k), and one group multiplied kmargin0 of different k values with fixed coefficient 0.3. The experimental results are shown in table 4-5. It can be seen that comparing the kmargin0 of different k values multiplied by different coefficients (-k) with the fixed coefficient 0.3 can significantly reduce the number of iterations (15 times reduced to 12 times), and obtain the performance improvement of Rank-1 and mAP by 2.13\% and 2.85\% respectively. As can be seen from table 4-5, compared with SRD, ARD can reduce the number of iterations and significantly improve the performance, which proves the superiority of ARD algorithm.

\begin{table}[]
\begin{center}
\begin{tabular}{l|l|lll}
\hline
No. & Method         & \multicolumn{3}{c}{DukeMTMC-VID}                                      \\ \cline{3-5}
    &                & \multicolumn{1}{l|}{Total Steps} & \multicolumn{1}{l|}{R-1}   & mAP   \\ \hline
1   & IDE+EUG        & \multicolumn{1}{l|}{20}          & \multicolumn{1}{l|}{72.79} & 63.23 \\
2   & PAM+SRD        & \multicolumn{1}{l|}{16}          & \multicolumn{1}{l|}{83.33} & 77.13 \\
3   & PAM+ARD(0.3)   & \multicolumn{1}{l|}{15}          & \multicolumn{1}{l|}{87.04} & 82.31 \\
4   & PAM+ARD(-k) & \multicolumn{1}{l|}{12}          & \multicolumn{1}{l|}{89.17} & 85.16 \\ \hline
\end{tabular}
\end{center}
\caption{The ablation study to verify the effectiveness of the dynamic coefficient for ARD. We replace the dynamic coefficient with static number 0.3.}
\end{table}

\subsection{Analysis and Visualization.}
We visualize the relationship between the accuracy and the number of samples of the adaptive sampling ARD (Algorithm 3) on the DukeMTMC-VID data set with the number of iterations of the algorithm, as shown in Figure 4-6:

It can be clearly seen that the increasing trend of the number of samples is only gentle when k=1.0, and stable rising trend when k= 0.7, 0.8 and 0.9. Therefore, learning in each iteration is meaningful. MAP accuracy curve shows an obvious rise in the form of four steps, with each value of k corresponding to a step, and the accuracy of iteration has been increasing. When the final algorithm converges, the accuracy rate is basically the highest, indicating that the performance of the last iteration model is the best, and there is no need to select the optimal model through additional verification sets.

Experiment 4, 5 sample along with the iteration number of absolute value change as shown in figure 4, can see more clearly for different kmargin0 multiplied by the coefficient of different k value (1.2-k) play a role in iteration algorithm: make the sampling number increasing trend is more stable and fast, thus reducing the total number of iterations, improve the learning efficiency of the algorithm.

\begin{figure}[t]
\begin{center}
\includegraphics[width=0.9\linewidth]{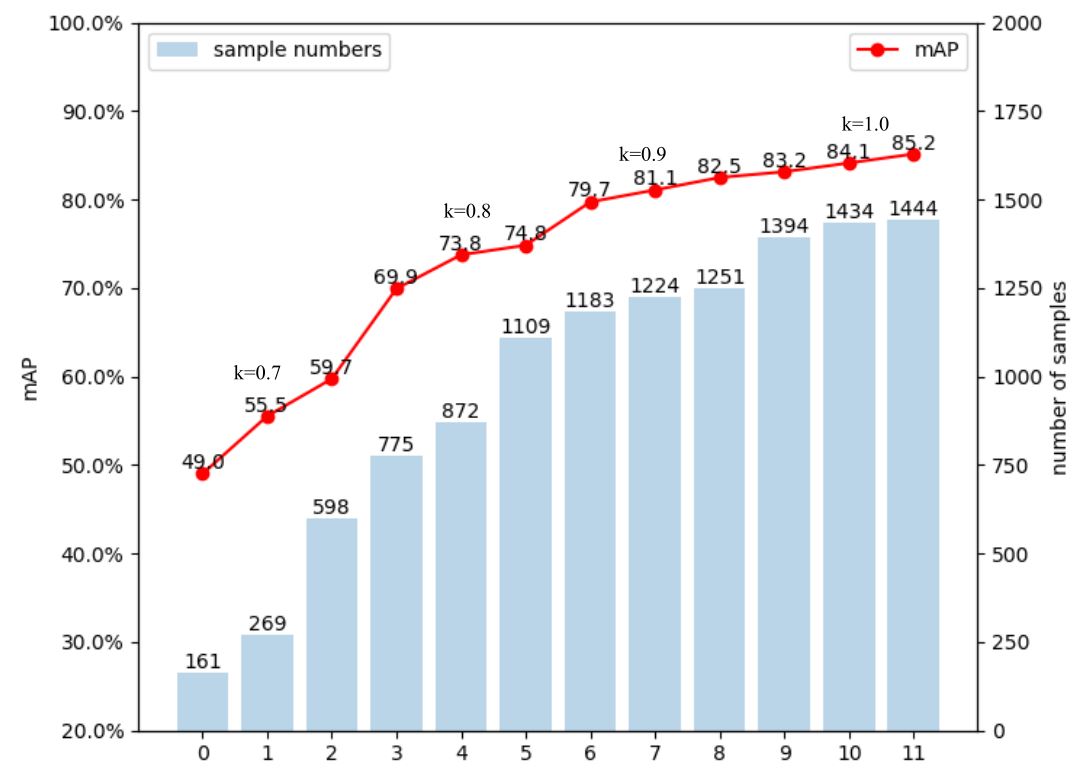}
\end{center}
   \caption{The increasing trend of sampling numbers and mAP along with iteration steps on DukeMTMC. We can clearly see 4 step-by-step growth with different k.}
\label{fig:long}
\label{fig:onecol}
\end{figure}

\begin{figure}[t]
\begin{center}
\includegraphics[width=0.9\linewidth]{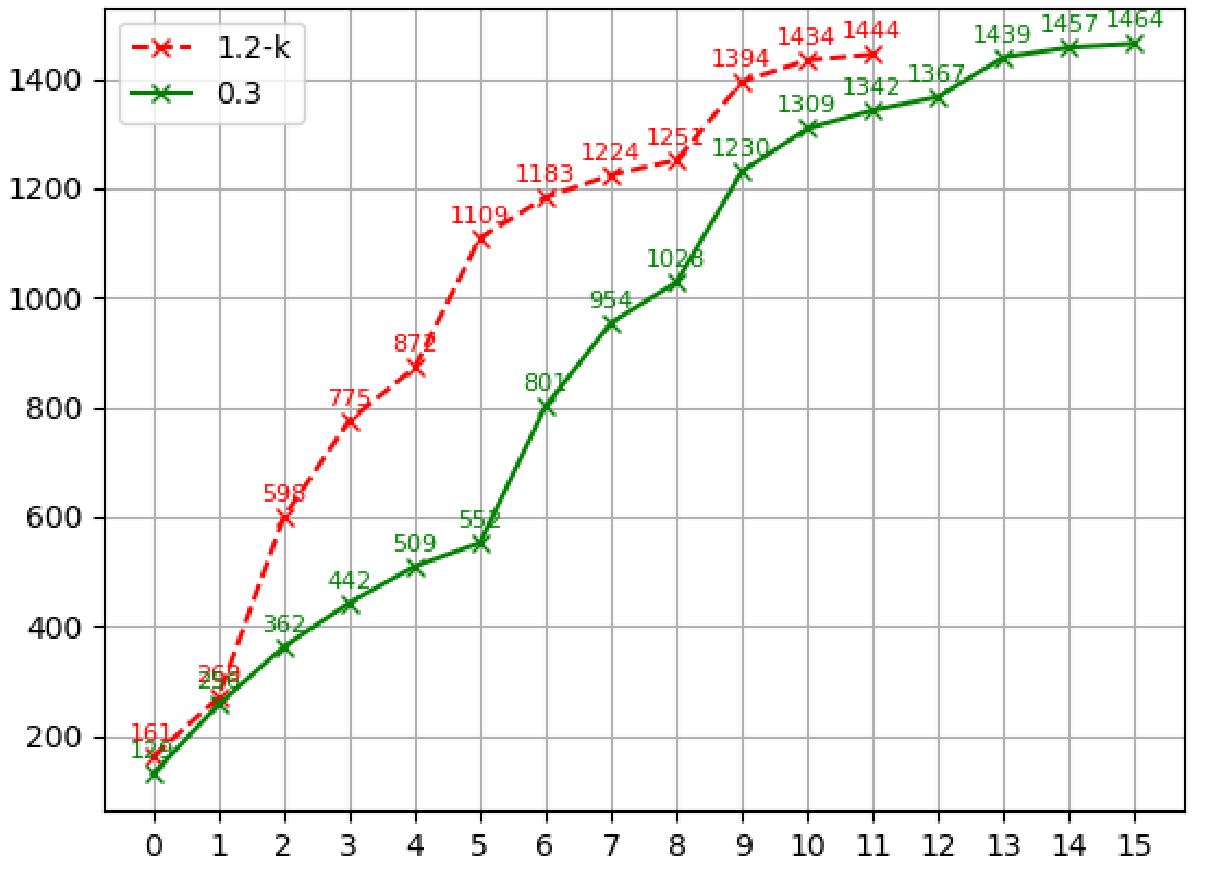}
\end{center}
   \caption{Sampling number of dynamic coefficient and static coefficient increases at different paces along with iteration steps on DukeMTMC. It can be seen that dynamic coefficient grows more fastly.}
\label{fig:long}
\label{fig:onecol}
\end{figure}

\section{Conclusion}
Since unlabeled person tracklets are cheap and easy to get, data driven deep models can get promising results with label estimation for few-example person re-identification. The challenge is that how to estimate labels correctly and select the reliable ones to enlarge traing set. In the paper, we propose a light and converges fast network PAM, which is suitable for few-example person re-ID. We also propose an adaptively sampling strategy to select most reliable pseudo label samples and gradually learn a more robust model. Our approach surpasses the state-of-the-art method by 5.5 19.0 16.4 points (absolute) in rank-1 accuracy on PRID2011 iLIDS-VID DukeMTMC-VID, and 2.9 points in mAP (absolute) on MAS. The proposed approach is very efficient and accurate for few-example person re-identification.
%-------------------------------------------------------------------------

{\small
\bibliographystyle{ieee_fullname}
\bibliography{egbib}
}

\end{document}